\documentclass{article}
\usepackage{arxiv}
\usepackage[utf8]{inputenc} 
\usepackage[T1]{fontenc}    
\usepackage{url}            
\usepackage{booktabs}       
\usepackage{amsfonts}       
\usepackage{nicefrac}       
\usepackage{microtype}      
\usepackage{graphicx}
\usepackage{tcolorbox}
\usepackage{multirow}
\usepackage{diagbox}
\usepackage[normalem]{ulem}
\usepackage{doi}
\usepackage{placeins} 
\usepackage{enumitem}
\usepackage{amsmath} 
\usepackage{xspace}
\usepackage{siunitx}
\usepackage{authblk}
\usepackage{titling}
\usepackage{subcaption}
\usepackage[numbers]{natbib}

\usepackage{hyperref}       

\newcommand{\sysname}{Ling-Coder-Lite\xspace}

\title{Every Sample Matters: Leveraging Mixture-of-Experts and High-Quality Data for Efficient and Accurate Code LLM}

\author{CodeFuse \& Ling Team, Ant Group}

\begin{document}

\maketitle

\thispagestyle{firstpage}

\begin{abstract}
Recent advancements in code large language models (LLMs) have demonstrated remarkable capabilities in code generation and understanding. 
It is still challenging to build a code LLM with comprehensive performance yet ultimate efficiency.  Many attempts have been released in the open source community to break the trade-off between performance and efficiency, such as the Qwen Coder series and the DeepSeek Coder series. 
This paper introduces yet another attempt in this area, namely Ling-Coder-Lite.
We leverage the efficient Mixture-of-Experts (MoE) architecture along with a set of high-quality data curation methods (especially those based on program analytics) to build an efficient yet powerful code LLM. 
Ling-Coder-Lite exhibits on-par performance on 12 representative coding benchmarks compared to state-of-the-art models of similar size, such as Qwen2.5-Coder-7B and DeepSeek-Coder-V2-Lite, while offering competitive latency and throughput. In practice, we achieve a 50\% reduction in deployment resources compared to the similar-sized dense model without performance loss. To facilitate further research and development in this area, we open-source our models as well as a substantial portion of high-quality data for the annealing and post-training stages. The models and data can be accessed at~\url{https://huggingface.co/inclusionAI/Ling-Coder-lite}.

\end{abstract}
\begin{figure}[ht]
    \centering
    \begin{subfigure}{0.14\textwidth}
        \centering
        \includegraphics[width=\linewidth,height=3.3\linewidth]{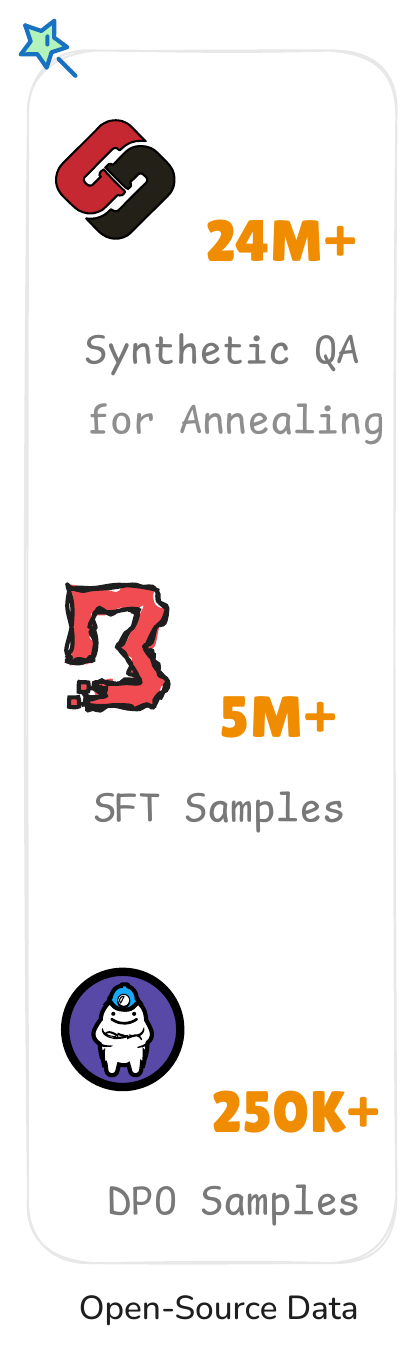}
        \caption{}
        \label{fig:merge_op_data}
    \end{subfigure}
    \begin{subfigure}{0.5\textwidth}
        \centering
        \includegraphics[width=\linewidth]{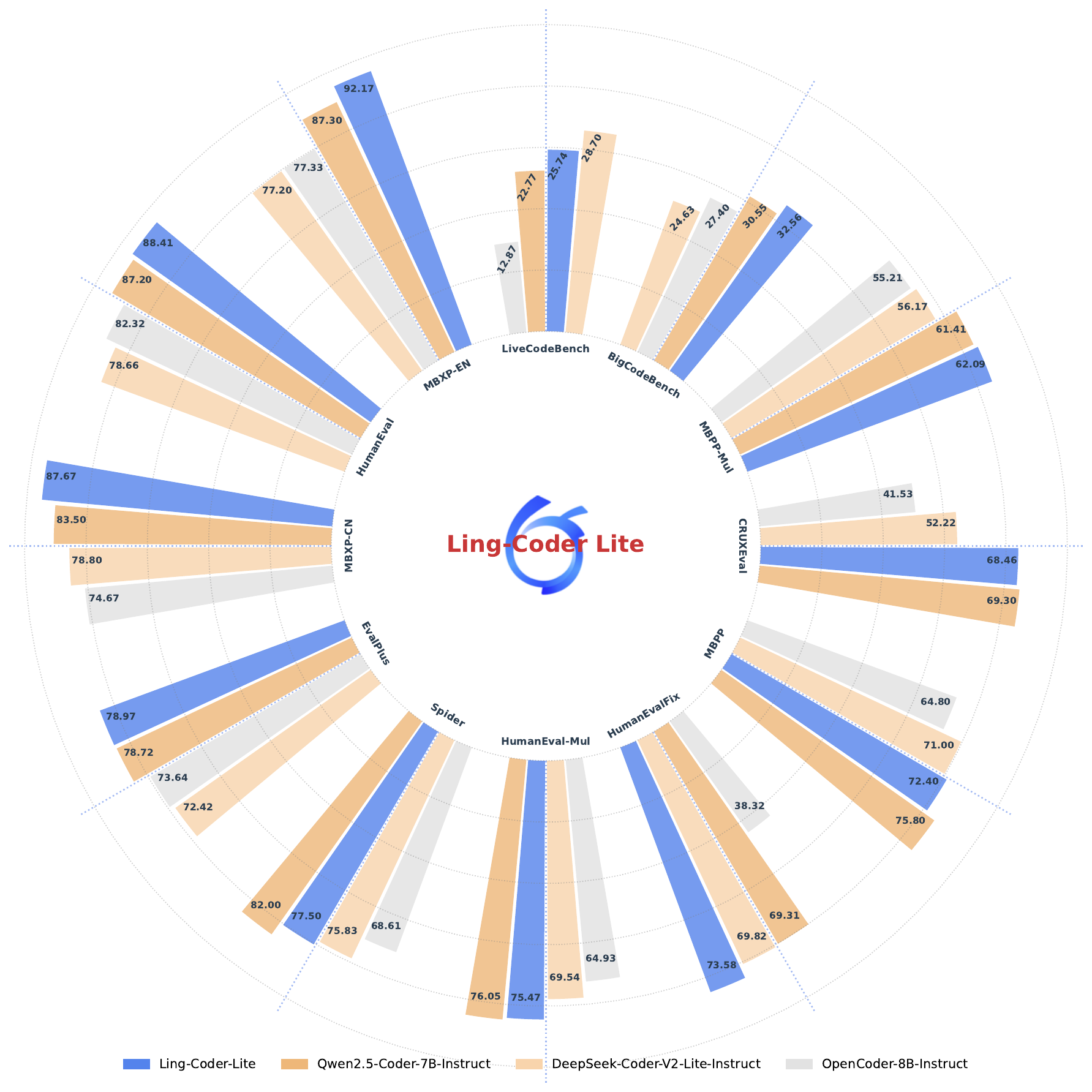}
        \caption{}
        \label{fig:merge_performance}
    \end{subfigure}
    \begin{subfigure}{0.35\textwidth}
        \centering
        \includegraphics[width=\linewidth,height=1.25\linewidth]{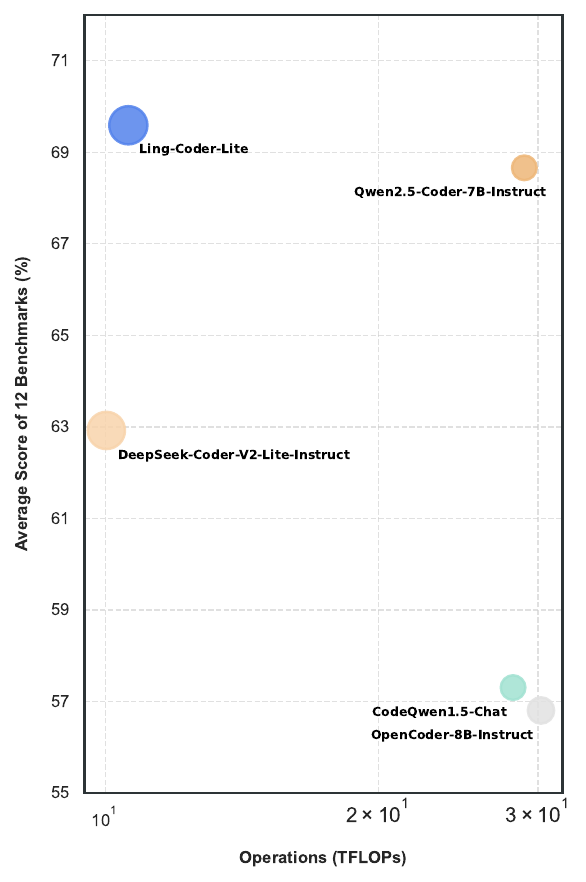}
        \caption{}
        \label{fig:merge_accu_tflops}
    \end{subfigure}
    \caption{\sysname achieves an effective trade-off between high performance and efficiency by leveraging high-quality data. (a) A substantial portion of high-quality data used in the \sysname training process (approximately 30 million samples) has been released as open-source data; (b) Average performance of various code LLMs with similar parameter size on 12 code benchmarks; (c) A comparison of various models over performance (in terms of average evaluation scores) versus the theoretical number of computational operations (in terms of TFLOPs per single inference with a context length of 4096).}
    \label{fig:merge}
\end{figure}

\section{Introduction}

Code LLMs, such as the StarCoder series~\citep{li2023starcoder, lozhkov2024starcoder}, CodeLlama~\citep{roziere2023code}, Qwen-Coder series~\citep{qwen,hui2024qwen2}, DeepSeek-Coder series~\citep{guo2024deepseekcoderlargelanguagemodel,zhu2024deepseek}, CodeStral~\citep{codestral2024}, and OpenCoder~\citep{huang2024opencoder}, have made significant advancements thanks to the efforts of the open source community. The code capabilities of state-of-the-art open-source code LLMs, such as DeepSeek-Coder-V2 (236B)~\citep{zhu2024deepseek} and Qwen2.5-Coder-32B~\citep{hui2024qwen2}, are now very close to, or even surpass, the best proprietary models available at the time of their release.
However, larger models such as Qwen2.5-Coder-32B deliver powerful performance but relatively low efficiency.
It is still quite challenging to build a code LLM that achieves both powerful performance and ultimate efficiency.
Given the importance of inference efficiency in practical scenarios like code completion in AI-IDE,
we introduce \sysname, which employs an MoE architecture with a total of $16.8$B parameters but only $2.75$B activated parameters. Compared to other high-performing small models of similar parameter sizes, such as DeepSeek-Coder-V2-Lite (16B)~\citep{zhu2024deepseek}, Qwen2.5-Coder-7B~\citep{hui2024qwen2}, and OpenCoder-8B~\citep{huang2024opencoder}, \sysname demonstrates on-par overall performance on a bunch of code evaluation benchmarks, while achieving competitive latency and throughput.

\sysname is developed by continuously training over an intermediate checkpoint of the Ling-Lite-Base model~\citep{lingteam2025flopcountsscaling300b}.
The initial checkpoint is trained on natural language dominated corpus with about 7T tokens, while Ling-Coder-Lite-Base is further trained with an additional 3.2T tokens corpus dominating with code data, culminating in a total of over 10T tokens. The instruction model \sysname is then obtained from Ling-Coder-Lite-Base via post-training alignment.
The continuous training consists of several stages, with each stage involving varying ratios of code data, math data, and natural language data.
The performance improvement is attributed to not only the well-designed training policy but also a set of high-quality data.
We acquire around {1.4T} code data, comprising about {836B} source code data, {119B} code-related text data, {177B} synthetic code data, as well as {182B} high-quality filtered code data and {132B} repo-level code data derived from original code corpus. Note that \textbf{a portion of high-quality synthetic data is open-sourced on HuggingFace~\footnote{\url{https://huggingface.co/datasets/inclusionAI/Ling-Coder-SyntheticQA}}}.
For mathematical data, similar to the approaches used in DeepSeekMath~\citep{shao2024deepseekmath}, we recall {200B} math-related data from Common Crawl.
For natural language data, we directly sample from the Ling-Lite-Base model's training corpus.
All of these data undergo a rigorous curation pipeline developed on our big data platform before being approved for use.

During the post-training phase, we enhance our model through a two-stage process. First, we conduct supervised fine-tuning (SFT)~\citep{ouyang2022training,liu2024mftcoder,gong2024coba} using millions of curated code samples, combined with Ling-Lite~\citep{lingteam2025flopcountsscaling300b} existing math and natural language SFT data. Second, we apply the Direct Preference Optimization (DPO) algorithm~\citep{rafailov2024directpreferenceoptimizationlanguage} to align the model with human preferences. This alignment phase incorporates hundreds of thousands of code-specific preference samples, alongside Ling-Lite~\citep{lingteam2025flopcountsscaling300b} specialized math and natural language preference data. These steps significantly improve \sysname's coding capabilities while ensuring better alignment with human preferences. \textbf{Notably, we open-source a substantial portion of our SFT and DPO code data on HuggingFace~\footnote{\url{https://huggingface.co/datasets/inclusionAI/Ling-Coder-SFT}}~\footnote{\url{https://huggingface.co/datasets/inclusionAI/Ling-Coder-DPO}}}. {Finally, we develop and open-source \sysname~\footnote{\url{https://huggingface.co/inclusionAI/Ling-Coder-lite}}, an efficient yet high-performance code LLM.}
We extensively evaluate our model on 12 representative code benchmarks compared to state-of-the-art (SOTA) small code LLMs of similar size, such as Qwen2.5-Coder-7B, DeepSeek-Coder-V2-Lite, and OpenCoder-8B.
Overall results are illustrated in Figure~\ref{fig:merge_performance}, showing that \sysname achieves overall on-par performance (wins in 7 out of 12 benchmarks) to these small SOTA code LLMs.

The main contributions can be summarized as follows:
\begin{itemize}[leftmargin=*]
    \item We introduce and open-source the Ling-Coder-Lite and Ling-Coder-Lite-Base models, based on the Ling-MoE architecture~\citep{lingteam2025flopcountsscaling300b}, featuring 16.8B total parameters with only 2.75B activated during inference. They are licensed for both research and commercial usages under the MIT license, and are particularly suitable in scenarios demanding low-latency responses, such as code completion tasks in the AI-IDE.

    \item We present details on how to build a competitive code LLM, including techniques for high-quality code-related text data curation and question-answering (QA) data synthesis, and training recipes especially for data mixing and stage scheduling details.
    We have publicly released a substantial portion of the dataset used to train the Ling-Coder-Lite model, including a subset of synthetic  QA data for annealing phase, a subset of post-training data, totaling approximately 30 million samples as shown in Figure~\ref{fig:merge_op_data}.

    \item We conduct extensive evaluations across {12} representative code benchmarks. The results demonstrate that \sysname exhibits on-par overall performance (wins in 7 out of 12 benchmarks) in comparison to state-of-the-art open-source code models with similar parameter sizes, such as Qwen2.5-Coder-7B and DeepSeek-Coder-V2-Lite, while offering competitive efficiency as shown in Figure~\ref{fig:merge_accu_tflops}. In practice, we achieve 50\% deployment resources reduction compared to similar-sized dense models without performance loss.
\end{itemize}

\section{Data Collection and Curation}
\label{sec:data_collection}

\sysname is continuously trained on 3.2T tokens starting from a {7T} intermediate checkpoint obtained from Ling-Lite-Base~\citep{lingteam2025flopcountsscaling300b}.
Throughout several stages of continuous training with different data mixing, the training corpus is dominated by code data, accounting for at least 60\%.
Since the natural language and math corpora are directly sampled from the training data of the Ling-Lite foundation, our focus lies on the code data.

The code data comprises raw source code (at both file and repository levels), code-related text, and synthesized QA data.
To gather this data, we primarily utilize publicly available raw data from sources such as GitHub, Common Crawl~\footnote{https://commoncrawl.org}, and GH Archive~\footnote{\label{footnote_gharchive}https://www.gharchive.org}, supplemented by code data from open-source datasets including The Stack~\citep{Kocetkov2022TheStack}, the Stack v2~\citep{lozhkov2024starcoder}, Dolma~\citep{dolma}, and Matrix~\citep{zhang2024mapneo}. In the following sections, we will elaborate on our data preprocessing pipeline and the specific methodologies employed in constructing each type of data.

\subsection{Data Curation Pipeline} \label{preprocessing-pipeline}

\begin{figure}[h]
    \centering    \includegraphics[width=0.9\textwidth]{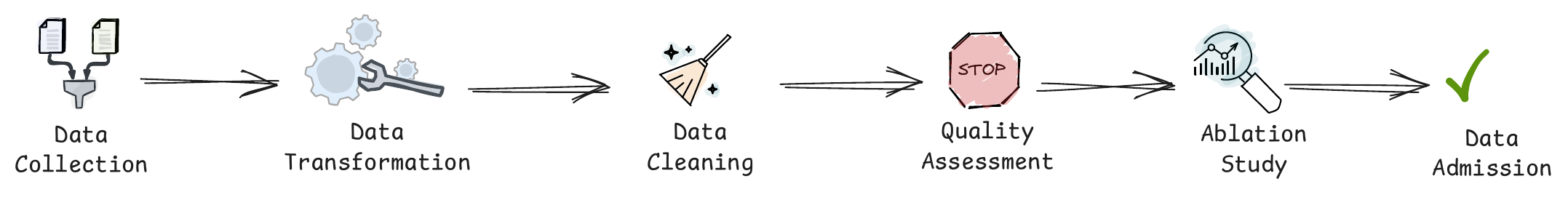}
    \caption{Data Curation Pipeline for Code Data Utilized by~\sysname.}
    \label{fig:preprocess_pipeline}
\end{figure}

We establish a standardized data curation pipeline utilizing our big data platform, as illustrated in Figure~\ref{fig:preprocess_pipeline}, which encompasses data collection, transformation, cleaning, quality inspection, and ablation study. Each dataset utilized in the pre-training of \sysname is subjected to this pipeline to ensure its eligibility for inclusion.

In the data collection phase, we gather public repositories from GitHub as of {June 2024} and Common Crawl data as of {August 2024}. Besides, we supplement these with open-source datasets such as the Stack~\citep{Kocetkov2022TheStack,lozhkov2024starcoder} and Matrix~\citep{zhang2024mapneo} corpora.

In the data transformation phase, we filter, assemble, and convert the raw data into formats suitable for LLM training. For instance, we convert the Common Crawl data from HTML pages into plain text, and selectively retrieve code-related content.

Subsequently, the processed data undergoes a cleansing phase, which includes rule-based filtering, deduplication, and security sanitization.
We develop two distinct sets of filtering rules: one for source code and the other for code-related text data.
For source code, we adopt StarCoder~\citep{li2023starcoder,lozhkov2024starcoder}'s filtering criteria, as employed in DeepSeek-Coder~\citep{guo2024deepseekcoderlargelanguagemodel,zhu2024deepseek}, and enhance them with additional rules based on quality assessment feedback. For instance, we exclude samples where URLs or IPv4/6 addresses exceed 60\% of the content, samples where email addresses, phone numbers, or date-time strings make up more than 50\%, samples containing garbled characters, and samples with rates of inter-line repetition or word duplication exceeding 70\%.
For code-related textual data, such as code-related webpages, markdown files, notebook data, pull requests, issues, and synthesized QA data, we develop a separate set of filtering rules, since the source code filtering rules cannot be directly reused, especially those limiting maximum line length to 1,000 characters or average line length to 100 characters.
There exist new rules specific to code-related text data, such as rules to remove samples that contain image references, links, or placeholders.
After applying the filtering rules, we further process the resulting data with the near-deduplication method~\citep{Kocetkov2022TheStack} and remove samples containing toxic content.

In the quality assessment phase, the processed data undergoes full rule-based verification, LLM-based sample scoring, and manual sample scoring. Data receiving a composite quality score below a threshold (e.g., 85 out of 100) is deemed low quality and sent back for reprocessing. Data meeting the quality threshold advances to the next phase.

In the ablation study phase, we conduct experiments using a 1B small model to compare performance between a control group and an experimental group.
The control group continues training with the original dataset for an additional amount of tokens (e.g., 50B). The experimental group mixes the studied code data with the sampled original dataset at a 1:1 ratio, and trains for the same number of tokens from the same checkpoint. The new dataset is approved only if the experimental group achieves better performance than the control group across various code evaluation benchmarks.

\subsection{Source Code}

Leveraging the curation pipeline as illustrated in Figure~\ref{fig:preprocess_pipeline}, we obtain a total of \num{1100}B tokens of source code data across {618} programming languages, comprising 836B tokens of file-level data, 182B tokens of filtered high-quality data, and 132B tokens of repository-level data.
In the following, we will describe the detailed methods on how to construct high-quality source code data and repository-level data.

\subsubsection{High-Quality Code Filtering}

To further collect high-quality source code data, we design multiple filter operators to filter low-quality data. These filters primarily operate by leveraging code metrics, code quality score, and repository metadata.

The code metrics filter evaluates several key aspects, including comment ratio, effective lines of code, and syntax correctness as determined through abstract syntax tree analysis. For code quality scoring, we implement a classification model based on FastText \citep{joulin2016fasttext} trained on high-quality code data labeled by a teacher LLM. Repository metadata filters consider factors such as star count, fork count, and the number of files within each repository.

By applying these comprehensive filters, we successfully identify high-quality file-level code data characterized by elevated comment ratios and superior quality scores. This filtered dataset comprises approximately 95B tokens, which are subsequently utilized for annealing training. This rigorous selection process ensures that our model is trained on exemplary coding practices and patterns.
As an independent step, we utilize BERT classifiers, perplexity distribution analysis, askLLM~\citep{sachdeva2024traindataefficientllms}, and other techniques to filter another high-quality code corpus, totaling about 87B.

\subsubsection{Repository-Level Data Constructing}

Traditional code LLM training processes source code at the file level, neglecting inter-file dependencies within a project. Previous studies ~\citep{chen2021codex,li2023starcoder} emphasize that this approach fails to capture the structural relationships in real-world codebases, hindering models from handling project-level code. To address this, we implement a topological sorting algorithm that analyzes and leverages file dependencies within a repository by identifying import relationships and ordering files so that dependencies precede dependent files. Our approach advances beyond the regex-based import identification method from DeepSeek-Coder~\citep{guo2024deepseekcoderlargelanguagemodel}  by employing a more sophisticated AST (Abstract Syntax Tree)-based analysis, executed in a two-phase process.

\textbf{Phase 1: Constructing the Import Dependency Graph.}
This phase involves four key steps. First, we implement a virtual file system to process repository data without restoring the directory structure, significantly reducing disk I/O operations. Second, we utilize Tree-sitter~\footnote{https://tree-sitter.github.io/tree-sitter} to generate ASTs for supported languages, enabling precise identification of import statements at the AST level instead of using regex patterns. Third, for each import statement, we determine the corresponding module file within the repository through path-based searching and package naming convention analysis, combining both methods for complex cases. Finally, we construct a directed graph where nodes represent files and edges denote import dependencies.

\textbf{Phase 2: Lexicographical Topological Sorting.}
Our sorting algorithm extends lexicographical topological sort to accommodate both DAGs (Directed Acyclic Graphs) and cyclic graphs, ensuring a unique, stable ordering. The process begins with input validation, confirming that the graph is directed. Next, each node is assigned a unique ID for consistent processing, followed by calculating the in-degree, i.e., the number of incoming edges, for each node. During iterative processing, nodes with the minimum in-degree are identified and prioritized using lexicographical order in case of ties. A node is selected to "break" cycles if the minimum in-degree is greater than zero, which is then removed from the graph. The in-degrees of its successors are updated, and these steps are repeated until the graph is empty.

The algorithm produces a total ordering where for each edge $(u, v)$ in the original graph, $u$ appears before $v$ in the result, subject to lexicographical ordering when multiple valid orderings exist.

We aggregate file-level code data, curated with the pipeline as shown in Figure~\ref{fig:preprocess_pipeline}, by repository. Each file within a repository is then sorted and concatenated into a complete sample using the proposed repository concatenation method. It is important to note that our concatenation algorithm supports 16 programming languages but cannot analyze implicit import dependencies or dependencies from dynamic code calls. As a result, some repositories cannot be successfully concatenated. For those that are successful cases, we conduct a high-quality filtering process based on several repository attributes, including the average code quality score per file (achieved using a FastText model), average comment ratio, and average number of effective code lines. In the end, we obtain 132B repository-level code data.

\subsection{Code-Related Data}

We also process and obtain 119B code-related textual data, including code-related web pages, code-comment pairs, notebooks, Markdown files, GitHub Pull Requests (PRs), and GitHub issues.

\subsubsection{Code-Related Webpages}
Similar to DeepSeek-Coder-V2~\citep{zhu2024deepseek}, DeepSeekMath~\citep{shao2024deepseekmath}, Qwen2.5-Coder~\citep{hui2024qwen2}, and OpenCoder~\citep{huang2024opencoder},  we also develop a pipeline to recall code-related web pages from over 280 billion pages in Common Crawl, resulting in a 66B dataset.

\begin{figure}[!h]
    \centering
    \includegraphics[width=0.80\textwidth]{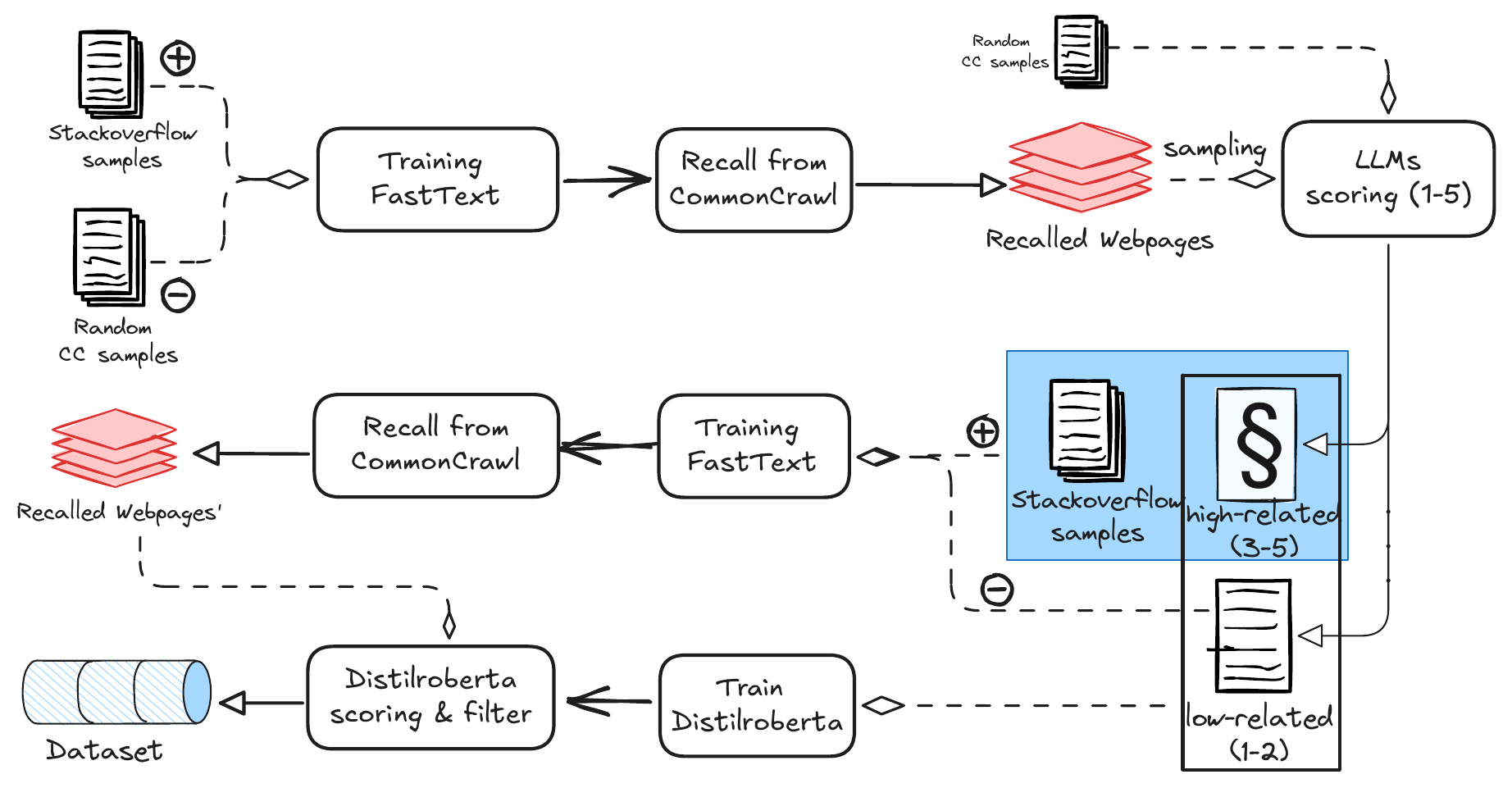}
    \caption{Pipeline for Recalling Code-Related Data from Common Crawl.}
    \label{fig:cc_recall_pipeline}
\end{figure}

Our recall process, illustrated in Figure~\ref{fig:cc_recall_pipeline}, comprises three stages. In the first stage, following DeepSeek-Coder-V2's methodology, we train a FastText~\citep{joulin2016bag} classifier using 500K StackOverflow samples as positive instances and 500K randomly sampled Common Crawl pages (parsed from raw HTML to plain text) as negative instances. Notably, we modify the FastText library~\footnote{https://github.com/facebookresearch/fastText} source code to support BPE (Byte-Pair Encoding) tokenization, aligning with DeepSeek-Coder-V2's implementation. This classifier is then deployed on our big data platform to recall code-related web pages from 280B Common Crawl pages. Initial quality inspection reveals suboptimal data quality from this first-round retrieval.
For the second stage, we diverge from DeepSeek-Coder-V2 by employing an askLLM~\citep{sachdeva2024traindataefficientllms} strategy to enhance training data quality. Specifically, we leverage open-source LLMs to score randomly sampled Common Crawl pages and initial recalled results across three dimensions: code relevance, educational value, and data cleanliness (1-5 scale). Samples with score $\ge3$ (combined with selected StackOverflow samples) form the positive set (about 1 million), while samples with scores $<3$ constitute the negative set (about 1 million). Despite retraining FastText model with this refined data, quality inspection still indicates insufficient improvement.
In the final stage, we implement a DistilRoberta~\citep{Sanh2019DistilBERTAD}-based filtering system. Using LLM-generated scores from the second stage, we fine-tune a scoring model on the DistilRoberta-base~\footnote{https://huggingface.co/distilbert/distilroberta-base} model. This model evaluates second-stage recalled results, ultimately selecting only pages with maximum scores (5/5) for our final dataset, yielding 66B data.

Notably, we find that the quality of parsing raw Common Crawl pages to plain text affects our results significantly. To address this, we implement targeted improvements in parsing quality specifically for mathematical and code content.

\subsubsection{Code-Comment Pairs}
We obtain a set of high-quality code-comment pairs (\emph{COCO}s for short) by crawling about 1.46B COCOs from the Stack v2 and other popular GitHub repositories with at least 10 stars.
To ensure the quality of the data, we conduct a two-phase sanitization process following our prior attempts on code change understanding~\citep{li2024hqcm}: (1) Rule-based sanitization; (2) Distillation-based sanitization.
The first phase involves removing syntactically low-quality COCOs, while the second phase focuses on ensuring their semantic quality.

\begin{itemize}[leftmargin=*]
    \item \textbf{Rule-based sanitization.}
    Syntactic sanitization is carried out using seven heuristic rules (i -- vii) commonly found in existing literature~\citep{panth21deepjit,codexglue,di2023codefuse13b}.
    The initial step involves removing duplicate COCOs based on their {MD5 value} (i).
    Subsequently, we set constraints on the code and comment of each COCO.
    For the code, we require each COCO to consist of {30} to {100,000} characters (ii) and {1} to {100} lines (iii).
    As for the comment, each COCO should be {non-empty} (iv), with {30} to {100,000} characters (v) and {less than 80\% special characters} (vi), e.g., \texttt{/}, \texttt{@}, \texttt{*}, and \texttt{\#}.
    Finally, comments with {special annotations} (vii), such as \texttt{TODO:}, \texttt{BUG:}, and \texttt{FIXME:}, are excluded as these are commonly used by developers to convey information unrelated to the code's semantics.

    \item \textbf{Distillation-based sanitization.}
    Semantic sanitization ensures that the comment of each COCO is in line with the semantics (e.g., behavior, intention, functionality, etc.) of its code.
    As we did not find suitable existing tools for this purpose, we conduct the process by training a specialized small model using distilled data from a teacher LLM; this is a popular and effective technique~\citep{xu2024distillsurvey,rong25c4rllama}.
    Initially, we prompt the teacher LLM to evaluate the semantic consistency between the code and comment of 25.6K selected COCOs step by step.
    Specifically, for each COCO, we instruct it to:
    (a) Break down the provided code line by line, detailing what each line accomplishes;
    (b) Summarize the detailed, line-by-line analyses into one concise sentence capturing the code's essence;
    (c) Review the provided comment, pinpointing its central message, analyzing its consistency with the summary from Steps a \& b, and providing a clear, step-by-step rationale for the evaluation;
    (d) Conclude the evaluation by reaching an overall consistency (true or false) between the provided code and comment.
    Following this, we fine-tune a small language model, Qwen2-0.5B, using the labeled 25.6K COCOs, with 19.2K used for training and the remainder equally split for testing and validation.%
    \footnote{The fine-tuned model achieves 0.9021 precision, 0.9198 recall, and 0.9109 F1 on the distilled data.}
    In order to enhance the reasoning capability of Qwen2-0.5B, our fine-tuning process is auto-regressive, with the goal of enabling the small model to learn the step-by-step thought chain.
    Finally, we utilize the fine-tuned model to reason and label the remaining data; we remove those that are negatively labeled.
\end{itemize}

As a result, our sanitization process yields  $\sim$17B tokens, spanning 18 popular programming languages including Java, JavaScript, Python, C++, Shell, and Lua among others.

\subsubsection{Miscellaneous Code-related Data}

\textbf{Pull Request.} Following StarCoder2 \citep{lozhkov2024starcoder}, we adopt an analogous methodology for constructing Pull Request (PR) data. Initially, textual data from PRs is downloaded from the GH Archive~\textsuperscript{\ref{footnote_gharchive}}.
Subsequently, commit information is extracted from these textual data. Following this, associated commit diff code data is scraped from GitHub based on the extracted commit information. Thereafter, we employ the same data cleansing strategies as delineated in StarCoder2 to sanitize these code data. Finally, the textual and code data belonging to a single Pull Request is concatenated following the approach detailed in StarCoder2. A deviation in our PR filtering strategy from that of StarCoder2 lies in the exclusion of PRs with more than three base files, while ensuring all integral files are retained as context for the respective PR. Our processing encompasses all Pull Requests on GitHub within the timeframe of 2015 to 2018, culminating in a corpus of 1.1 million PRs totaling 8B tokens.

\textbf{Notebook.} We also construct Script data based on notebook data by referencing StarCoder2's methodology. We crawl all notebook data from GitHub, and after construction and cleaning, we ultimately obtain 14B data. Our ablation experiments indicate that this data significantly enhances the model's capabilities.
We also collect and process open-source notebook data from The Stack~\citep{Kocetkov2022TheStack}, as well as notebook data from the Kaggle~\footnote{https://www.kaggle.com}, resulting in extra 3B data.
We speculate that this is due to the inherent characteristics of notebook data. Compared to typical code data, notebook data contains richer textual information, with code blocks often serving independent functions and contexts usually being confined within a single file. These characteristics may assist the model in learning the close connections between text and code.

\textbf{Markdown.}
Similar to the StarCoder series~\citep{li2023starcoder,lozhkov2024starcoder} and DeepSeek-Coder-V2~\citep{zhu2024deepseek}, we also treat Markdown files collected from GitHub as a category of code-related data. Given that the raw Markdown files from GitHub often contain non-code-related content or even toxic content, we opt to use the DistilRoberta scoring model, trained for recalling code-related web data from Common Crawl, to first evaluate the code relevance and educational value. Only those samples with high code relevance and educational value are subjected to subsequent cleaning, quality inspection, and ablation study processes. In the end, we obtain approximately 11B Markdown data.

\subsection{Synthetic Data}
\label{sec:synthetic_data}
With the rapid advancements in model training and the swift consumption of publicly available internet data, synthetic data has become increasingly important and feasible.
In the training of \sysname, we utilize two types of synthetic data. The first type is generated using the SOTA open-source LLMs with the Magpie~\citep{xu2024magpie} method, which is used during the pre-training stage, totaling approximately 163B tokens.
The second type consists of post-training synthetic data that is generated using methods similar to OSS-Instruct~\citep{wei2023magicoder} and Evol-Instruct~\citep{xu2023wizardlm}, comprising over 10 million samples. Portions of this data, along with other moderate-quality SFT data, are utilized as pre-training data during the annealing phase, altogether amounting to about 14B.

\subsubsection{Synthesis with Magpie}

Magpie~\citep{xu2024magpie} is a data synthesis pipeline designed to produce high-quality alignment data without relying on prompt engineering or seed questions. It directly constructs instruction data by using a pre-query template to prompt aligned LLMs for sampling instructions. Typically, the input to these aligned LLMs comprises a pre-query template, query content, and a post-query template.
For instance, the Llama2-Chat model accepts input in the format "[INST] Hi! [/INST]," where "[INST]" serves as the pre-query template and "[/INST]" as the post-query template.
When only the pre-query template is provided, the LLM autoregressively generates the query content; once the full input is assembled, the LLM's response can be obtained. The generation process involves two main steps:

\begin{itemize}[leftmargin=*]

    \item \textbf{Instruction Generation.} Magpie constructs an input query based on the LLM's predefined instruction template. This query specifies the role of the instruction provider (e.g., user) without detailing any specific instruction. The auto-regressive LLM has been fine-tuned on instruction data formatted according to this template, allowing it to autonomously generate instructions when fed the Magpie-crafted query.

    \item \textbf{Response Generation.} Magpie transmits the generated instruction to the LLM, which then processes it to produce the corresponding responses.
\end{itemize}

For a specific teacher LLM, we employ the Magpie technique to synthesize instruction data covering multiple coding scenarios. These scenarios include text-to-code for multiple programming languages, test case generation, code explanation, code repair, code refactoring, and code execution prediction,
all aiming at enhancing the coding capabilities of \sysname.
Through this process, we ultimately synthesize approximately 0.15 billion cleaned and detoxified question-answer pairs, about 163B tokens, which are subsequently used during annealing. This synthetic dataset plays a crucial role in improving the model's ability to understand and generate code across diverse programming contexts.

\textbf{Practical Considerations.} For the instruction generation process, it is advisable to set relatively high values for both \textit{TEMPERATURE} and \textit{TOP-P} to promote diversity in the generated instructions. Conversely, during response generation, using greedy decoding or a lower temperature is recommended, as the high-probability tokens are likely derived from the LLM's training data.

\subsubsection{Instruction Data Synthesis.}\label{sec:sftdata}

Drawing from the methods of OSS-Instruct~\citep{wei2023magicoder} and Evol-Instruct~\citep{xu2023wizardlm}, we make slight modifications and combine them for synthesizing instruction data. Some of the synthesized instruction data is used as pre-training data, while another portion is allocated for fine-tuning. Specifically, we propose two synthesis approaches: Bottom-Up synthesis and Top-Down synthesis.

\begin{figure}[h]
    \centering    \includegraphics[width=0.95\textwidth]{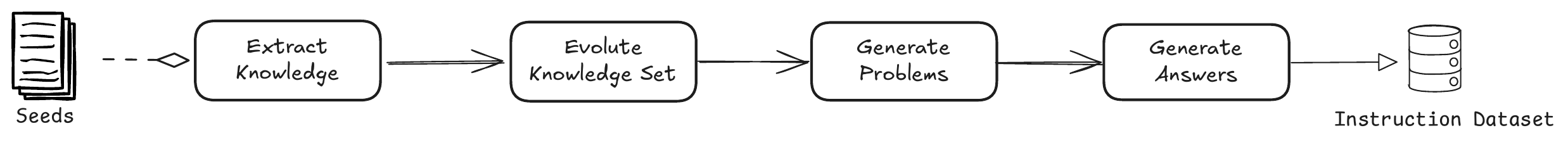}
    \caption{Pipeline of Bottom-Up Instruction Data Synthesis.}
    \label{fig:bottom_up}
\end{figure}

\textbf{Bottom-Up Synthesis.}
Figure~\ref{fig:bottom_up} illustrates the synthesis process of the bottom-up approach. We start by collecting a set of seeds (code pieces and code-related text), similar to the work~\citep{huang2024key}.
In the first stage, we use a state-of-the-art LLM to extract hierarchical knowledge from each seed sample, encompassing high-level topics and their corresponding detailed key points.
In the second stage, inspired by Evol-Instruct \citep{xu2023wizardlm}, we expand the knowledge set by prompting the LLM to generate five evolved variations for each piece of knowledge, ensuring these variations are related to—but distinct from—the original.
In the third stage, we generate problems based on both the evolved knowledge. Applying techniques from Arena~\citep{li2024crowdsourced}, we convert interrogative sentences to declarative ones and generate seven unique problems for each piece of knowledge.
As a result, the original seed problems are expanded
$(K+1)\times N$ times. In our experiments, we set
$K = 7$ and $N = 5$, acknowledging that larger values may lead to redundancy. Ultimately, this process generates more than 10 million instruction samples.

All synthesized instruction data, whether used for fine-tuning or pre-training, first undergoes a standard data cleaning pipeline for SFT data, which includes rule-based anomaly detection, cleansing, detoxification, quality inspection, ablation study, etc.

\textbf{Top-Down Synthesis.}
In the bottom-up synthesis approach, the knowledge of synthetic problems is limited by seed problems, even with knowledge evolution. To enhance the diversity of synthetic data, we propose a top-down synthesis pipeline that leverages the knowledge from a wide range of coding books.
We utilize an LLM to generate content based on approximately 200 book titles related to popular programming languages (e.g., Python) and their libraries (e.g., PyTorch). Instead of reproducing entire books, we instruct the LLM to create a hierarchical knowledge structure. This process consists of two steps:

\begin{itemize}[leftmargin=*]
\item The LLM generates a catalog for each book title, structured in three levels—chapter, section, and subsection—forming a tree.

\item Next, the LLM provides fine-grained knowledge for each leaf node (section or subsection), specifying its higher-level node to differentiate between similar sections across different books (e.g., "Conditional Statements" in both Python and C books).
\end{itemize}

Subtopics represent finer levels of detail within Knowledge Points, with the LLM deciding their necessity. To improve the distinctiveness of generated problems, we provide the path from each leaf node to the root node in the format: "the \textit{[LEAF]} that belongs to the section \textit{[SECTION]} that belongs to the chapter \textit{[CHAPTER]} that belongs to the book \textit{[BOOK]}.”

After completing these steps, we obtain a comprehensive knowledge tree for each book. We extract all knowledge paths from subtopics to the book name and prompt the LLM to create coding problems based on these paths. The LLM can also incorporate additional relevant topics to create more challenging problems. Ultimately, we synthesize around 1.2 million problems and then use the LLMs to generate answers for each problem.

\section{Model Architecture and Training Details}

\subsection{Model Architecture}
Ling-Coder-Lite is continuously trained over the Ling-Lite-Base~\citep{lingteam2025flopcountsscaling300b}, which is Mixture of Experts (MoE) architecture with 16.8B total parameters and 2.75B activated parameters. Ling-Coder-Lite is configured with 28 transformer layers and a hidden dimension of 2048. Each MoE layer comprises 2 shared experts and 64 routed experts, with an intermediate hidden dimension of 1408 per expert. During the routing process, each token activates 6 out of the 64 routed experts. To ensure efficient and stable training, we followed the same fine-grained expert and NormHead strategy as introduced in Ling-Lite~\citep{lingteam2025flopcountsscaling300b}.

\subsection{Training Details}

\begin{figure}[h]
    \centering
    \includegraphics[width=0.95\textwidth]{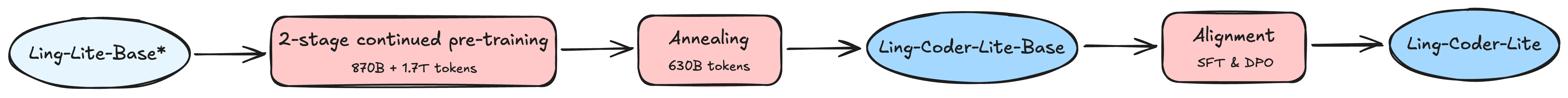}
    \caption{Training Pipeline for Ling-Coder-Lite.}
    \label{fig:training_pipeline}
\end{figure}

Figure~\ref{fig:training_pipeline} shows the overall training pipeline of Ling-Coder-Lite. It is noteworthy that we start the continuous pretraining from an intermediate 7T-token checkpoint of Ling-Lite (denoted as Ling-Lite-Base$^{*}$), while Ling-Lite finally stopped the initial pre-training phase with 9T tokens~\citep{lingteam2025flopcountsscaling300b}.
We conduct continuous pre-training from Ling-Lite-Base$^{*}$ with objectives including Next-Token-Prediction and Fill-In-Middle (FIM). In the final stage of pre-training, we introduce an annealing phase similar to \citep{touvron2023llama2}, in which we enhance training data quality and employ a learning rate annealing strategy to capture more refined features. Subsequently, we apply Supervised Fine-tuning (SFT) and Direct Preference Optimization (DPO) to significantly improve the practicality and reliability for code-related tasks.

\begin{table}[t]
    \centering
    \caption{Pre-training Setting of Each Stage.}
    \setlength{\tabcolsep}{5pt}
    \resizebox{\linewidth}{!}{
    \begin{tabular}{lcccc|ccc|ccc}
       \toprule
       \multirow{2}{*}{\pmb{Stage}} & \multirow{2}{*}{\pmb{LR}} & \multirow{2}{*}{\pmb{Scheduler}} & \multirow{2}{*}{\pmb{Batch Size}} & \multirow{2}{*}{\pmb{Tokens}} & \multicolumn{3}{c|}{\pmb{Token Ratio}} & \multicolumn{3}{c}{\pmb{Token Ratio in Code Data}} \\
        & & & & & Code & NLP & Math & Raw Code & Code-Related & Synthetic \\
       \midrule
       Stage 1 & 3e-4 $\sim$ 1.5e-4 & multi-step & 16M & 870B & 70 & 20 & 10 & 94 & 5 & 1 \\
       Stage 2 & 1.4e-4 $\sim$ 1.1e-4 & cosine & 16M & 1.7T & 65 & 20 & 15 & 75 & 20 & 5 \\
       Annealing & 1.4e-4 $\sim$ 1.4e-6 & inverse square root & 16M & 630B & 60 & 20 & 20 & 40 & 10 & 50 \\
       \bottomrule
    \end{tabular}}
    \label{tab:pre-training-setting}
\end{table}

\subsubsection{Continuous Pre-Training}
Due to constraints in computational resources and data production iteration speed, we conduct a two-phase continued pre-training process prior to annealing. Table~\ref{tab:pre-training-setting} presents the training parameters and data distribution settings for our pre-training phases. Overall, we gradually reduce the proportion of code, particularly raw code, as we observe that an excessive ratio of this component tends to compromise natural language understanding capabilities.

\textbf{Stage 1.} This stage involves training on 870B tokens with a batch size of 16M tokens. The maximum and minimum learning rates are set at 3e-4 and 1.5e-4, respectively. We implement a multi-step constant learning rate scheduler strategy. Specifically, the learning rate remains constant within each step, with no decay. When the training loss reaches a plateau or the validation loss stops decreasing for several consecutive iterations, the learning rate is further reduced to promote continued convergence.

\textbf{Stage 2.} In this stage, we train the model on 1.7T tokens with a batch size of 16M tokens. The learning rate is configured with a maximum of 1.4e-4 and a minimum of 1.1e-4, employing a cosine decay schedule to promote sustained convergence of the training loss. We strategically adjust the data composition, reducing the proportion of Raw Code while increasing the ratio of Code-Related and Synthetic data. This strategic adjustment aims to enhance the model's comprehension of more complex coding tasks, and further improve its performance.

\subsubsection{Annealing}
The final stage of pre-training involves 630B tokens, maintaining a consistent batch size of 16M tokens. We implement an inverse square root decay scheduler for learning rate annealing, rapidly reducing the learning rate from 1.4e-4 to 1.4e-6. Concurrently, we further adjust the data composition by decreasing the proportion of Raw Code while substantially increasing the ratio of Synthetic data (being meticulously constructed and of the highest quality). Our strategy leverages the annealing stage for more refined and stable absorption of high-quality knowledge and significantly enhances performance.

\subsubsection{Supervised Fine-tuning}
\textbf{Data Preparation.}
In SFT stage, we collect 18.8 million instruction data, including 11.7 million code-related instructions. This dataset includes not only code and mathematical instructions but also a subset of pure natural language instruction data from Ling-Lite~\citep{lingteam2025flopcountsscaling300b}, and the distribution is approximately 3:1:1 for code, general natural language, and mathematical instructions, respectively. This inclusion aims to maintain the model's performance on general tasks while simultaneously enhancing its ability to follow instructions for complex code-related problem-solving.

In addition, we synthesize a large set of instructional data targeting specific code-related problem domains. We begin by identifying several key domains such as text-to-code, reasoning, SQL, and coding competitions. We then ask state-of-the-art (SOTA) models to generate problem seeds using a combination of top-down and bottom-up approaches as described in Section~\ref{sec:synthetic_data}. With the help of Evol-Instruct~\citep{xu2023wizardlm} and Oss-Instruct~\citep{wei2023magicoder}, we further expand upon these seeds, creating additional problems. Then we utilize SOTA code LLMs to generate answers for these problems. Finally, the resulting data undergo a rigorous deduplication decontamination and quality assessment process, as outlined in Section~\ref{sec:data_collection}.

\textbf{Training.} Ling-Coder-Lite is fine-tuned with 5 epochs over the 18.8 million samples dataset. We implement a cosine schedule with 100-step warm-up. The initial learning rate is set at 4e-5, and the total batch size is configured at 384. To optimize training efficiency, we employ a data packing technique~\citep{liu2024mftcoder}.

\subsubsection{Direct Preference Optimization}
\textbf{Data Preparation.} We begin by assembling a comprehensive collection of open-source DPO datasets.
To supplement these existing datasets, we generate additional DPO datasets following the OSS-Instruct methodology. This process involves two steps: First, we use a teacher LLM to generate code-related questions based on randomly sampled code snippets from the Stack~\citep{Kocetkov2022TheStack}. Then, we generate multiple answers for each question using various language models, including Ling-Coder-Lite-SFT, and other SOTA LLMs.

For quality assurance, we process both the open-source and synthesized datasets using the Llama-3.1-Nemotron-70B-Reward-HF~\citep{grattafiori2024llama} reward model. This model ranks all answers for each question, and we filter out data points where the highest reward score falls below a predetermined threshold. Through ablation studies, we demonstrate that this filtering process is crucial for enhancing the overall quality of the DPO data.

\textbf{Training.} Following the development of the SFT model, we implement offline Direct Preference Optimization (DPO) to further align Ling-Coder-Lite with human preferences. The DPO process involves training 2 epochs over 1.4 million chosen-rejected pairing samples. We set the learning rate at 3e-7 and the total batch size at 192 samples, and we use a cosine schedule with 150-step warm-up. Our empirical observations indicate that DPO significantly enhances the model's capability in open-ended generation tasks without substantive changes in performance across standard benchmarks.

\section{Evaluation}

To rigorously evaluate the performance of Ling-Coder-Lite models, we conduct comprehensive multi-dimensional evaluations within the standardized CodeFuseEval framework~\footnote{\url{https://github.com/codefuse-ai/codefuse-evaluation}}, ensuring reproducibility and comparability of results
\footnote{Note that we did not directly take result numbers from papers of those compared models, since for a portion of benchmarks different models had slightly different benchmark configurations which makes their result numbers can't be directly comparable. For instance, for the MBPP benchmarks, some may report results in 3-shot, while others report results in 0-shot. 
Some some report results on 500 samples, while others report results on 1000 samples.}. 
The evaluation is divided into two parts for base models and instruct models.

\subsection{Evaluation Configurations}
For code generation tasks, we employ the pass@k metric, while Text-to-SQL tasks are evaluated using execution accuracy (EX), with a consistent greedy decoding strategy applied across all tasks and a maximum output length constraint of 4,096 tokens. The prompt configuration prioritizes adherence to the original setups of the evaluated open-source models, defaulting to their respective benchmark settings when model-specific configurations are unavailable. This approach ensures technical precision while aligning with established academic and operational conventions.

\subsection{Benchmarks}

To thoroughly evaluate the \sysname and Ling-Coder-Lite-Base's capabilities, we collect and evaluate their performance on 12 representative code-related benchmarks, categorized into 6 task types, as detailed below:

\begin{itemize}[leftmargin=*]
 \item \textbf{HumanEval}~\citep{chen2021codex} is a widely recognized dataset for evaluating Python code completion, comprising 164 carefully curated problems created by researchers at OpenAI. It emphasizes the fundamental Python programming capabilities of LLMs and falls under the category of \textit{code generation} tasks.

 \item \textbf{MBPP}~\citep{austin2021program} comprises 1000 Python programming problems, constructed through crowdsourcing, primarily targeting a model's proficiency in basic Python. In our evaluation, we select 500 problems with ID 11-510 from MBPP (i.e. MBPP-500) to evaluate the text-to-code generation capability, specifically generating code based on problem descriptions. It falls under the \textit{code generation} task category.

 \item \textbf{EvalPlus}~\citep{evalplus} is an enhanced version of HumanEval and MBPP, referred to as HumanEval+ and MBPP+, respectively. 
 HumanEval+ extend the test-cases of the HumanEval benchmark by 80x, while MBPP+ selects 378 problems from MBPP-500 and extend the test-cases by 35x. 
 EvalPlus provides a more accurate evaluation of the model's coding capabilities and also falls under the \textit{code generation} task category.

 \item \textbf{LiveCodeBench (LCB)}~\citep{jain2024livecodebench} collects problems from periodic contests on LeetCode, AtCoder, and Codeforces platforms and uses them for constructing a holistic benchmark for evaluating Code LLMs across variety of code-related scenarios continuously over time. 
 It consists of four subsets: self-repair, code execution, test output prediction, and code generation. In this study, we focus solely on its \textit{Advanced Code Generation} subset, using evaluation data from August 2024 to November 2024, which comprises 22 easy, 34 medium, and 45 hard problems.
 It falls under the \textit{advanced code generation} task category.

 \item \textbf{BigCodeBench (BCB)}~\citep{zhuo2024bigcodebench} aims to evaluate the true programming capabilities of LLMs in a more realistic setting. It is designed for HumanEval-like function-level code generation tasks, but with much more complex instructions and diverse function calls. 
 It consists of two splits, \textsc{Complete} and \textsc{Instruct}, each containing \num{1140} problems. We use the Complete split for testing base models and the Instruct split for testing chat models.
It falls under the \textit{advance code generation} task category.

 \item \textbf{MultiPL-E} extends the HumanEval and MBPP benchmarks to 18 languages to evaluate the multi-language performance of code LLMs. In our evaluations, following the language choices of the Qwen2.5-Coder model, we select 10 most widely used programming languages: Python, Java, JavaScript, TypeScript, C++, C\#, PHP, Rust, Go, and Bash. To align with the evaluations of the Qwen2.5-Coder model, we divide this dataset into two versions: MultiPL-E HumanEval (abbr. MultiPLE-H) and MultiPL-E MBPP (abbr. MultiPLE-M). It falls under the \textit{multi-language code generation} task category.
 
 \item \textbf{MBXP-PLUS} bilingual evaluation benchmark is an optimized version of the open-source MBXP~\citep{mbxp_athiwaratkun2022}, consisting of two components: \textbf{MBXP-EN} (the English version) and \textbf{MBXP-CN} (the Chinese version). It focuses on the task of natural language to code generation. This benchmark covers six programming languages: Java, C++, JavaScript, TypeScript, Go, and PHP, providing a total of 1,200 task instances, with 100 instances per language. Each task instance includes a canonical solution, ensuring a 100\% self-validation pass rate (pass@1). Furthermore, this dataset systematically evaluates the effectiveness of prompts and test cases, using pass@k as its primary metric.
 It falls under the \textit{multi-language code generation} task category.

 \item   \textbf{HumanEvalFix}~\citep{muennighoff2023octopack} is a benchmarking suite constructed from HumanEval and its translations in five additional languages, designed to assess the code repair capabilities of large models. It is divided into two parts: the primary focus of the paper is on the first part, where only test cases are provided without docstrings; the second part provides docstrings as the source of truth instead of test cases. It falls under the \textit{code fixing} task category.

 \item \textbf{CRUXEval}~\citep{gu2024cruxeval} (Code Reasoning, Understanding, and eXecution Evaluation) is a benchmark of 800 Python functions and input-output pairs. The benchmark comprises two tasks: CRUXEval-I, which involves predicting the input from the output, and CRUXEval-O, which involves predicting the output from the input. It falls under the \textit{code understanding and reasoning} task category

\item \textbf{DS-1000}~\citep{Lai2022DS1000} focuses on assessing a model's ability to perform data science analysis using Python code, covering essential libraries such as Numpy, Pandas, TensorFlow, Pytorch, Scipy, Sklearn, and Matplotlib. It falls under the \textit{data analysis application} task category

\item \textbf{Spider} is a large human-labeled dataset for complex and cross-domain semantic parsing and text-to-SQL benchmark containing 2,147 evaluation samples. It also falls under the \textit{data analysis application} task category

\end{itemize}

\subsection{Evaluation on Base Model}

\begin{table}[t]
  \caption{Performance of various models on HumanEval, MBPP and BigCodeBench (Greedy Mode, Pass@1).} 
  \centering
  \resizebox{0.95\linewidth}{!}{
  \begin{tabular}{l|cc|cc|cc|cc|c}
  \toprule
   \multirow{2}{*}{\textbf{Model}} & \multicolumn{2}{c|}{\textbf{Size}} & \multicolumn{2}{c|}{\textbf{HumanEval}} & \multicolumn{2}{c|}{\textbf{MBPP}} & \multicolumn{2}{c|}{\textbf{BigCodeBench}} & \multirow{2}{*}{\textbf{Average}} \\
    & \#TP & \#AP & HE & HE+ & 3-shot & MBPP+ & Full & Hard & \\
  \midrule
  \hline
  
  DS-Coder-V2-Lite-Base & 16B & 2.4B & 40.9 & 34.1 & 62.6 & 59.4 & 30.6 & 8.1 & 39.3 \\
  
  Qwen2.5-Coder-7B-Base & 7B & -- &  61.6 & 53.0 & 68.8 & 62.9 & 45.8 & 16.2 & 51.4 \\
  
  OpenCoder-8B-Base & 8B & -- & 66.5 & 63.4 & 60.6 & 70.4 & 40.5 & 9.5 & 51.8 \\ 
  
  Ling-Coder-Lite-Base & 16.8B & 2.75B & 65.9 & 62.2 & 65.8 & 69.3 & 35.2 & 17.6 & \textbf{52.7} \\

  \hline
  \bottomrule
  \end{tabular}}
  \label{tab:base_model_evaluation_1}
\end{table}

\begin{table}[t]
  \caption{Performance of various models on CRUXEval (Greedy Mode).} 
  \centering
  \begin{tabular}{l|cc|cc|c}
  \toprule
   \multirow{2}{*}{\textbf{Model}} & \multicolumn{2}{c|}{\textbf{Size}} & \multicolumn{2}{c|}{\textbf{CruxEval}} & \multirow{2}{*}{\textbf{Average}} \\
    & \#TP & \#AP & Crux-I-CoT & Crux-O-CoT & \\
  \midrule
  \hline
  
  DS-Coder-V2-Lite-Base & 16B & 2.4B & 53.4 & 46.1 & 49.6 \\
  
  Qwen2.5-Coder-7B-Base & 7B & -- &  56.5 & 56.0 & 56.3 \\

  Ling-Coder-Lite-Base & 16.8B & 2.75B & 62.9 & 61.3 & \pmb{62.1} \\

  \hline
  \bottomrule
  \end{tabular}
  \label{tab:base_model_evaluation_2}
\end{table}

\begin{table}[t]
  \caption{Performance of various models on General and Mathematical benchmarks (Greedy Mode). We re-run the alignment test on all metrics. } 
  \centering
  \resizebox{\linewidth}{!}{
  \begin{tabular}{l|cc|cccccc|c}
  \toprule
   \multirow{2}{*}{\textbf{Model}} & \multicolumn{2}{c|}{\textbf{Size}} & \textbf{C-Eval} & \textbf{CMMLU} & \textbf{MMLU} & \textbf{BBH} & \textbf{GSM8K} & \textbf{MATH} & \multirow{2}{*}{\textbf{Average}} \\
    & \#TP & \#AP & \textit{5-shot} & \textit{5-shot} & \textit{5-shot} & \textit{3-shot} & \textit{4-shot} & \textit{4-shot} & \\
  \midrule
  \hline
  
  Ling-Lite-Base$^{*}$ (7T) & 16.8B & 2.75B & 62.2 & 62.5 & 60.0 & 51.5 & 49.1 & 18.1 & 50.6 \\
  
  DS-Coder-V2-Lite-Base & 16B & 2.4B & 62.3 & 63.1 & 60.0 & 67.1 & 66.8 & 32.3 & 58.6 \\
  
  Qwen2.5-Coder-7B-Base & 7B & -- & 69.1 & 72.7 & 70.5 & 67.3 & 83.4 & 42.2 & \pmb{67.5} \\

  Ling-Coder-Lite-Base & 16.8B & 2.75B & 74.8 & 75.5 & 67.6 & 69.3 & 78.8 & 40.5 & \pmb{67.7} \\

  \hline
  \bottomrule
  \end{tabular}}
  \label{tab:base_model_evaluation_3}
\end{table}

For the base model, we conduct a comprehensive comparison across three critical aspects: \textbf{Code Generation}, \textbf{Code Understanding and Reasoning}, and \textbf{General and Math}. Our comparative analysis focuses on the most popular and powerful open-source coder models, including \textbf{DeepSeek-Coder-V2-Lite-Base}, \textbf{Qwen2.5-Coder-7B-Base}, and \textbf{OpenCoder-8B-Base}. The performance metrics for these models are directly cited from their respective published papers, and we plan to supplement them with our own test results for these base models in the near future.

\subsubsection{Basic and Advanced Code Generation}
As shown in Table~\ref{tab:base_model_evaluation_1}, Ling-Coder-Lite-Base achieves the highest average score across all code generation tasks, achieving 52.7. This performance surpasses the second-best model OpenCoder-8B-Base by 0.9 points and exceeds DeepSeek-Coder-V2-Lite-Base by 13.4 points. These results indicate that Ling-Coder-Lite-Base demonstrates consistent strength across various code generation metrics without significant weaknesses.

\textbf{HumanEval.} To evaluate the foundational code snippet completion capabilities of the base models, we utilize both HumanEval and its extended version, HumanEval-Plus. As shown in Table~\ref{tab:base_model_evaluation_1}, the average of HumanEval is notably higher than both DS-Coder-V2-Lite-Base and Qwen2.5-Coder-7B-Base by at least 7 points, while only marginally trailing OpenCoder-8B-Base. This performance showcases Ling-Coder-Lite-Base's proficiency in basic code completion tasks.

\textbf{MBPP.} For assessing basic Text-to-Code generation capabilities, we employ the MBPP benchmark and its enhanced version, MBPP-Plus. Notably, the 3-shot version utilizes the MBPP-500 as test set. Ling-Coder-Lite-Base achieves an average score of 67.6 across both MBPP metrics, substantially leading the second-best Qwen2.5-Coder-7B-Base (65.9). This state-of-the-art performance among open-source coder models of similar size demonstrates Ling-Coder-Lite-Base's strong capability in understanding and solving fundamental coding problems.

\textbf{BigCodeBench.} To further evaluate the proficiency of the base models in complex instruction comprehension and diverse function invocation within practical programming scenarios, we introduce the BigCodeBench metric. As illustrated in Table~\ref{tab:base_model_evaluation_1}, Ling-Coder-Lite-Base's overall average BigCodeBench score is second only to Qwen2.5-Coder-7B-Base, while outperforming all models in the Hard subcategory.

\subsubsection{Code Understanding and Reasoning}
To assess whether the base models truly comprehend the logical reasoning and execution flow underlying code, we introduce the CRUXEval metric. It should be noted that the paper of OpenCoder-8B-Base does not report this metric, and we plan to supplement these results with our own evaluations in future versions of this report.

As illustrated in Table~\ref{tab:base_model_evaluation_2}, Ling-Coder-Lite-Base demonstrates impressive performance on both CRUXEval-I and CRUXEval-O, achieving scores of 62.9 and 61.3 respectively. The average score of CRUXEval surpasses all other models by at least 6 points, establishing a new state-of-the-art among open-source models of the same size. This remarkable performance can be attributed to our deliberate focus on scenarios involving code execution prediction when constructing synthetic data for training.

\subsubsection{General NLP and Mathematical Capabilities}
We greatly improve the coding performance for Ling-Coder-Lite-base while still maintaining the natural langue processing capabilities. To verify this, we evaluate the base model's capabilities in general natural language processing and mathematical reasoning, we utilize popular benchmarks including C-Eval~\citep{huang2023c}, CMMLU~\citep{li2306cmmlu}, MMLU~\citep{hendrycks2020measuring}, BigBench Hard (BBH)~\citep{suzgun2022challenging}, GSM8K~\citep{cobbe2021training}, and MATH~\citep{hendrycks2021measuring}. As shown in Table~\ref{tab:base_model_evaluation_3}, we re-run the alignment test on all metrics.
Ling-Coder-Lite-Base achieves an average score of 67.7 across these six metrics, outperforming DS-Coder-V2-Lite-Base and demonstrating comparable performance to Qwen2.5-Coder-7B-Base. While slightly underperforming Qwen2.5-Coder-7B-Base on mathematics-related metrics, Ling-Coder-Lite-Base exhibits marginally superior performance on other benchmarks, including C-Eval, CMMLU, and BBH. Notably, after extensive continued pre-training on code data, Ling-Coder-Lite-Base not only maintained but enhanced its general natural language capabilities. The model demonstrated an average improvement of 12 points on C-Eval, CMMLU, MMLU, and BBH metrics compared to Ling-Lite-Base$^{*}$. This improvement is attributed to maintaining at least 20\% of natural language text data during pre-training. Additionally, by incorporating 10\% to 20\% of mathematical data, the overall mathematical reasoning ability of Ling-Coder-Lite-Base improves by 25 points relative to Ling-Lite-Base$^{*}$, further enhancing the model's proficiency in mathematical reasoning.

\subsection{Evaluation on Instruct Model}

For the instruction model, we comprehensively evaluate the code capabilities of \sysname across five tasks—code generation, code reasoning, code fixing, data analysis applications, and SQL generation—utilizing a total of 12 benchmarks. 
We compare its performance with the previous state-of-the-art code LLMs of comparable parameter sizes, including DeepSeek-Coder-V2-Lite-Instruct~\citep{zhu2024deepseek}, Qwen2.5-Coder-7B-Instruct~\citep{hui2024qwen2}, and OpenCoder-8B-Instruct~\citep{huang2024opencoder}.
Notably, to ensure fairness, we re-evaluate all these models using the identical scripts and environments.

\subsubsection{Basic and Advanced Code Generation}

\textbf{HumanEval and MBPP benchmarks.} 
We evaluate the model's performance on HumanEval~\citep{chen2021codex}, MBPP~\citep{austin2021program}, and their enhanced version, EvalPlus~\citep{evalplus}, to evaluate its fundamental Python programming skills, as is commonly done with most LLMs.
For MBPP, we evaluate both its original version, which comprises a total of 500 tasks, and its enhanced version, MBPP+, which consists of 378 tasks. Both evaluations are conducted using a zero-shot approach.
As shown in Table~\ref{tab:humaneval_mbpp_eva}, \sysname outperforms all others on HumanEval and HumanEval+, exceeding the second-best model, Qwen2.5-Coder-7B-Instruct, by 1.21 and 3.22, respectively. 
On MBPP and MBPP+, \sysname ranked second, trailing the top-performing Qwen2.5-Coder-7B-Instruct by 3.4 and 3.16. 
Additionally, following the approach of the EvalPlus official leaderboard, we calculate the average scores for HumanEval+ and MBPP+. 
Numerically, \sysname emerges as the best-performing model, surpassing the second-best Qwen2.5-Coder-7B-Instruct by 0.25.
Given that the evaluation samples in EvalPlus feature more comprehensive test cases that better reflect a model's capabilities, there is reason to believe that \sysname performs slightly better than the previous the-state-of-the-art code LLMs of comparable parameter sizes.

\begin{table}[ht]
  \caption{Performance of \sysname and popular Open-Source Code LLMs of Comparable Parameter Size on HumanEval and MBPP Benchmarks (Greedy Mode, Pass@1)} 
  \centering
  \label{tab:humaneval_mbpp_eva}
  \resizebox{0.9\linewidth}{!}{
  \begin{tabular}{l|cc|cc|cc|c}
  \toprule
   \multirow{2}{*}{\textbf{Model}} & \multicolumn{2}{c|}{\textbf{Size}} & \multicolumn{2}{c|}{\textbf{HumanEval}} & \multicolumn{2}{c|}{\textbf{MBPP}} & \textbf{EvalPlus} \\
    & \#TP & \#AP & HumanEval & HumanEval+ & MBPP\tiny(500) & MBPP+ & avg. \\
  \midrule
  \hline
  CodeQwen1.5     &  7B & --  & 82.93   &  69.60   & 75.00 & 67.19 & 71.10 \\
  DeepSeek-Coder-V2 & 16B & 2.4B & 78.66 & 75.00 & 71.00 & 69.84 & 72.42 \\
  Qwen2.5-Coder     & 7B  & -- &  87.20 & 82.32 & \textbf{75.80} & \textbf{75.12} & 78.72 \\
  OpenCoder         & 8B  & -- & 82.32 & 77.44 & 64.80 & 69.84 & 73.64\\ 
  \sysname   & 16.8B  & 2.75B & \textbf{88.41} & \textbf{85.98} & 72.40 & 71.96 & \textbf{78.97} \\
  \hline
  \bottomrule
  \end{tabular}}
\end{table}

\textbf{Advanced Programming.}
Beyond basic Python programming skills, we also evaluate the model's ability to solve more challenging, competitive Python problems using benchmarks with higher complexity levels, specifically LiveCodeBench~\citep{jain2024livecodebench} and BigCodeBench~\citep{zhuo2024bigcodebench}.
For LiveCodeBench, we select problems from the period of August 2024 to November 2024, totaling 101 problems, which include 22 easy, 34 medium, and 45 hard problems. 
For BigCodeBench, we utilize the \textsc{Instruct} split, which is suitable for Instruct models and comprises 1,140 problems.
As shown in Table~\ref{tab:lcb_bcb_eva}, on the full LiveCodeBench set, \sysname ranks second, trailing DeepSeek-Coder-V2-lite-Instruct by 2.96, and the same holds true for the medium and hard subsets, where it outperforms both Qwen2.5-Coder-7B-Instruct and Opencoder-8B-Instruct models. 
On BigCodeBench, \sysname achieves the best performance on both the full set and the hard difficulty subset, surpassing the second-place Qwen2.5-Coder-7B-Instruct by 2.01 and 2.0, respectively. 
Overall, \sysname matches the current best models of comparable parameter sizes in tackling more complex, competitive high-level Python programming challenges.

\begin{table}[!t]
  \caption{Performance of \sysname and powerful Open-Source Code LLMs of Comparable Parameter Size on LiveCodeBench and BigCodeBench (Greedy Mode, Pass@1)} 
  \centering
  \label{tab:lcb_bcb_eva}
  \resizebox{0.9\linewidth}{!}{
  \begin{tabular}{l|cc|cccc|cc}
  \toprule
   \multirow{2}{*}{\textbf{Model}} & \multicolumn{2}{c|}{\textbf{Size}} & \multicolumn{4}{c|}{\textbf{LCB}\tiny{24.08-24.11}} & \multicolumn{2}{c}{\textbf{BCB}\tiny{-Instruct}} \\
    & \#TP & \#AP & Easy\tiny{(22)} & Medium\tiny{(34)} & Hard\tiny{(45)} & Full\tiny{(101)} & Hard & Full \\
  \midrule
  \hline
  CodeQwen1.5     &  7B & --  &  31.80  &  3.00   &  0.00 &  7.94 & 10.80 & 30.2 \\
  DeepSeek-Coder-V2 & 16B & 2.4B & 59.09 & \textbf{29.41} & \textbf{13.33} & \textbf{28.70} & 12.16 & 37.11\\
  Qwen2.5-Coder     & 7B  & -- & \textbf{63.60}  & 14.71 & 8.89 & 22.77 & 20.30 & 40.80 \\
  OpenCoder         & 8B  & -- & 45.45 & 2.94 & 4.44 &  12.87 & 12.16 & 32.72 \\ 
  \sysname   & 16.8B  & 2.75B & 54.55 & 26.47 & 11.11 & 25.74 & \textbf{22.3} & \textbf{42.81} \\
  \hline
  \bottomrule
  \end{tabular}}
\end{table}

\textbf{Multi-Language Programming.}
In addition to Python, we also test the model's performance on two multi-language benchmarks: \textbf{MultiPL-E} for code completion and \textbf{MBXP\_PLUS} for natural language to code. For MultiPL-E, we divide it into HumanEval and MBPP splits and select 10 popular programming languages: Python, C++, Java, PHP, TypeScript, C\#, Bash, Go, Rust, JavaScript. For MBXP\_PLUS, we create \textsc{English} and \textsc{Chinese} splits, with each split containing 100 problems across 6 programming languages: C++, Java, PHP, TypeScript, Go, Rust.

As shown in Table~\ref{tab:multiple_mbxp_eva}, on the HumanEval portion of MultiPL-E, \sysname is generally 0.58 behind the top-performing Qwen2.5-Coder-7B-Instruct model. 
Specifically, \sysname outperforms other models in Java and PHP, while it matches or falls short of Qwen2.5-Coder-7B-Instruct in other languages. 
On the MBPP portion of MultiPL-E, \sysname achieves the best overall performance, averaging 0.68 higher than the second-best Qwen2.5-Coder-7B-Instruct model. 
For specific languages, \sysname excels in C++, Java, PHP, C\#, Rust, and Bash, but underperforms compared to Qwen2.5-Coder-7B-Instruct in the other four languages. 
In the MBXP\_EN split, \sysname delivers the best overall performance, with a significant lead over the second-best model, except in PHP where it lags behind Qwen2.5-Coder-7B-Instruct. 
On the corresponding \textsc{Chinese} split, MBXP\_CN, \sysname shows the same leading advantage. 
Further observation reveals that all models perform significantly worse in Chinese across all programming languages compared to their English counterparts, suggesting a need to enhance Chinese training data. 
It's worth noting that during the evaluation, we find that \sysname occasionally answers a certain proportion of TypeScript questions using Python incorrectly. We plan to investigate and address this issue more thoroughly in our future research.
Overall, \sysname outperforms the other models in Table~\ref{tab:multiple_mbxp_eva}, demonstrating its powerful multi-language programming capabilities.

\begin{table}[!t]
  \caption{Performance of \sysname and Comparable Parameter Size Code LLMs on Multilingual Programming Tasks (Greedy Mode, Pass@1). TS=TypeScript; JS=JavaScript.} 
  \centering
  \label{tab:multiple_mbxp_eva}
  \scriptsize
  \resizebox{\linewidth}{!}{
  \begin{tabular}{l|cc|cccccccccc|c}
  \toprule
   \textbf{Model} & {\textbf{\#TP}} & \textbf{\#AP} & \textbf{Python} & \textbf{C++}  & \textbf{Java} & \textbf{PHP} & \textbf{TS} & \textbf{C\#} & \textbf{Bash} & \textbf{Go} & \textbf{Rust} & \textbf{JS}  & avg. \\
  \midrule
  \multicolumn{14}{c}{MultiPL-E \tiny{HumanEval}} \\
  \midrule
  \midrule
  CodeQwen1.5     &  7B & --  & 83.54 & 70.19 & 74.68 & 75.15 & 76.10 & 57.59 & 44.94 & 49.35 & 72.43 & 77.02 & 66.28\\
  DeepSeek-Coder-V2 & 16B & 2.4B & 79.27 & 70.19 & 71.52 & 72.05 & 80.50 & 69.62 & 41.77 & 58.44 & 70.51 & 81.37 & 69.54 \\
  Qwen2.5-Coder     & 7B  & -- & \textbf{85.98} & \textbf{75.78} & 77.22 & 75.16 & \textbf{81.13} & \textbf{78.48} & \textbf{51.90} & \textbf{76.62} & \textbf{75.64} & \textbf{82.61} & \textbf{76.05} \\
  OpenCoder         & 8B  & -- & 82.32 & 67.70 & 70.89 & 63.98 & 65.41 & 70.89 & 44.30 & 68.18 & 63.46 & {52.17} & 64.93\\ 
  \sysname   & 16.8B  & 2.75B & \textbf{85.98} & \textbf{75.78} & \textbf{82.28} & \textbf{78.88} & 80.50 & 77.22 & 49.37 & 72.08 & 73.72 & 78.88 & 75.47 \\
  \hline

  \multicolumn{14}{c}{MultiPL-E \tiny{MBPP}} \\
   \midrule
   \midrule
    CodeQwen1.5     &  7B & --  & 74.25 & 61.71 & 58.28 & 61.46 & 72.31 & 51.81 & \textbf{41.36} & 54.01 & \textbf{63.28}  & 67.51 & 60.6 \\
  DeepSeek-Coder-V2 & 16B & 2.4B & 75.00 & 62.40 & 63.21 & 59.95 & 43.59 & 46.11 & {28.27}  & 62.57 & 59.60 & 60.96 & 56.17 \\
  Qwen2.5-Coder     & 7B  & -- & \textbf{78.00} & 63.73 & 54.15 & 62.72 & \textbf{74.62}  & 51.04 & 33.25  & 63.37 & 62.43 & \textbf{70.78} & 61.41 \\
  OpenCoder         & 8B  & -- & 70.25 & 58.69 & 62.44 & 60.96 & 47.95 & 46.89 & 40.05  &  57.49 & 49.15 & 58.19  & 55.21 \\
  \sysname  & 16.8B  & 2.75B & 76.25 & \textbf{68.26} & \textbf{67.88} & \textbf{65.99} &  56.15 & \textbf{53.37} & 39.79  & \textbf{66.31} & 57.63 & 69.27 & \textbf{62.09} \\

  \midrule
 \multicolumn{14}{c}{MBXP \tiny{EN}} \\
 \midrule
 \midrule
 CodeQwen1.5     &  7B & --  & -- & 77.00 & 71.00 & 76.00 & 58.00 & --  & --  & 72.00 & -- & 90.00 & 73.80 \\
  DeepSeek-Coder-V2 & 16B & 2.4B & -- & 69.00 & 56.00 & 91.00 & 83.00 &  -- & -- & 73.00 & -- & 91.00 & 77.20 \\
  Qwen2.5-Coder     & 7B  & -- & -- & 87.00 & 90.00 & \textbf{95.00} & 78.00 & -- & -- & 85.00 & -- & 91.00 & 87.67 \\
  OpenCoder         & 8B  & -- & -- & 78.00 & 75.00 & 86.00 & 81.00 & -- & -- & 68.00 & -- & 76.00 & 77.33 \\
  \sysname   & 16.8B  & 2.75B & --  & \textbf{92.00}  & \textbf{92.00} & 92.00 & \textbf{91.00} & -- & -- & \textbf{89.00} & -- & \textbf{97.00} & \textbf{92.17} \\

  \midrule
 \multicolumn{14}{c}{MBXP \tiny{CN}} \\
 \midrule
 \midrule
 CodeQwen1.5     &  7B & --  & -- & 60.00 & 67.00 & 70.00 & 80.00 & --  & --  & 66.00 & -- &  81.00 & 70.70 \\
  DeepSeek-Coder-V2 & 16B & 2.4B & -- & 64.00 & 82.00 & 86.00 & 81.00 & -- & --  & 73.00 & -- & 87.00 & 78.80 \\
  Qwen2.5-Coder     & 7B  & -- & -- & 80.00 & 82.00 & \textbf{89.00} & 85.00 & --  & --  & 81.00 & -- & 84.00 & 83.50 \\
  OpenCoder         & 8B  & -- & -- & 71.00 & {73.00} & 78.00 & 79.00 & --  & --  & 74.00 & -- & 73.00 & 74.67 \\
  \sysname   & 16.8B  & 2.75B & -- & \textbf{86.00} & \textbf{90.00} & 88.00 & \textbf{87.00} & --  & --  & \textbf{82.00} & -- & \textbf{93.00} & \textbf{87.67} \\
  
  \bottomrule
  \end{tabular}}
\end{table}

\subsubsection{Code Understanding and Reasoning}

\textbf{CRUXEval.}
To evaluate our model's code comprehension and reasoning capabilities, we use the CRUXEval~\citep{gu2024cruxeval} benchmark, which is divided into two splits: CRUXEval-I and CRUXEval-O, each comprising 800 Python questions. In CRUXEval-I, the task is to provide a piece of Python code along with one of its execution output results and have the model deduce and infer an appropriate program input (i.e. arguments) that would produce this output. CRUXEval-O, on the other hand, presents the code and its input arguments, requiring the model to reason and predict the correct output. This evaluation approach allows a deeper examination of the model's understanding of Python code, including both forward and reverse execution logic.
During the evaluation, we employ the CoT (Chain of Thought) prompting method, guiding the model to first provide a step-by-step thought process before arriving at the final result. 

As shown in Table~\ref{tab:crux_eva}, \sysname performs best on the output prediction task CRUXEval-O, leading by a narrow margin of 0.42 over the second-place Qwen2.5-Coder-7B-Instruct. In CRUXEval-I, the rankings are reversed, with \sysname trailing by 2.1 but still significantly ahead of other models. Overall, \sysname is 0.84 behind the Qwen2.5-Coder-7B-Instruct model, yet holds a substantial lead over DeepSeek-Coder-V2-Instruct and OpenCoder-8B-Instruct, demonstrating powerful Python code comprehension and reasoning skills. Furthermore, we find that improvements in this evaluation metric also contribute to enhancing the model's code generation performance.

\begin{table}[!t]
  \caption{Performance of \sysname and Comparable Parameter Size Code LLMs on CRUXEval} 
  \centering
  \label{tab:crux_eva}
  \begin{tabular}{l|cc|ccc}
  \toprule
   \multirow{2}{*}{\textbf{Model}} & \multicolumn{2}{c|}{\textbf{Size}} & \multicolumn{3}{c}{\textbf{CruxEval}}  \\
    & \#TP & \#AP & Crux-I-CoT & Crux-O-CoT & Full  \\
  \midrule
  \hline
  CodeQwen1.5     &  7B & --  &  41.21  &  40.00  &  40.61  \\
  DeepSeek-Coder-V2 & 16B & 2.4B & 51.71 & 52.72 & 52.22 \\
  Qwen2.5-Coder     & 7B  & -- & \textbf{68.22}  & 70.38 &  \textbf{69.30}  \\
  OpenCoder         & 8B  & -- & 39.32 & 43.74 &  41.53   \\ 
  \sysname   & 16.8B  & 2.75B & 66.12 & \textbf{70.80} &  68.46  \\
  \hline
  \bottomrule
  \end{tabular}
\end{table}

\subsubsection{Code Fixing}


\begin{table}[!t]
  \caption{Performance of \sysname and Comparable Parameter Size Code LLMs on HumanEvalFix.} 
  \centering
  \label{tab:humanevalfix_eva}
  \resizebox{0.9\linewidth}{!}{
  \begin{tabular}{l|cc|ccccccc}
  \toprule
   \textbf{Model} & \textbf{\#TP} &\textbf{\#AP} & \textbf{Python}  
    & \textbf{Java} & \textbf{C++} & \textbf{Go} & \textbf{Rust} & \textbf{JavaScript} & \textbf{Overall}  \\
  \midrule
  \hline
  CodeQwen1.5     &  7B & --  &  48.78  &  56.10  &  45.73 & 48.17 & 43.90 & 50.00 & 48.78 \\
  DeepSeek-Coder-V2 & 16B & 2.4B & 71.95 & 60.37 & 63.41 & 71.95 & \textbf{67.07} & 76.83 & 68.60 \\
  Qwen2.5-Coder     & 7B  & -- & 78.65  & \textbf{79.26} &  49.39 & 76.22 & 54.87 & \textbf{77.44} & 69.31 \\
  OpenCoder         & 8B  & -- & 39.63 & 46.34 &  34.76  & {41.46} & {28.66} & {39.02} & {38.32}   \\ 
  \sysname   & 16.8B  & 2.75B & \textbf{83.54} & 76.83 &  \textbf{66.46} & \textbf{77.44} & 62.80 & 74.39 & \textbf{73.58} \\
  \hline
  \bottomrule
  \end{tabular}}
\end{table}

\textbf{HumanEvalFix.} We further evaluate the model's code repair capabilities using the HumanEvalFix~\citep{muennighoff2023octopack} benchmark. HumanEvalFix is built on the original HumanEval and translated versions in five additional languages: Java, C++, JavaScript, Go, and Rust. It consists of two splits: test and docstring. In our evaluation, we choose the test split, where test cases are provided without docstrings, requiring the model to identify and correct errors in the given buggy code based on the inputs and expected outputs described by the test cases. This evaluation assesses the model's ability to understand test cases and internal code logic, partially reflecting its capability to perform TTD (Test-Driven Development) tasks.
As shown in Table~\ref{tab:humanevalfix_eva}, \sysname performs best in Python, C++, and Go languages. For Java and JavaScript, Qwen2.5-Coder-7B-Instruct is the top performer, leading \sysname by 2.43 and 3.05, respectively. In Rust, the best performer is the DeepSeek-Coder-V2-Lite-Instruct model, followed by \sysname. Overall, \sysname has the best performance, with an average score 3.76 higher than the second-place DeepSeek-Coder-V2-Lite-Instruct model.

\subsubsection{Data Analysis Application}

We further evaluate the model's performance in data analysis applications, focusing on its ability to use popular Python data science-analysis libraries, as well as its capacity to generate SQL statements based on natural language descriptions.

\textbf{DS-1000.}
We use DS-1000~\citep{Lai2022DS1000} to evaluate our model's ability to solve practical data science analysis problems. The DS-1000 benchmark is a comprehensive dataset designed to evaluate and benchmark the capabilities of models in data science and analysis tasks. It encompasses a wide range of real-world data science problems, focusing on the practical application of data analysis, manipulation, and interpretation using Python.
Researchers collect and curated questions, reference answers, and test cases from StackOverflow, involving the use of seven Python data analysis libraries. These original questions form the basis of the benchmark. To avoid the possibility of the model memorizing these questions and answers from training data, researchers introduce three types of perturbations—Surface, Semantic, and Difficult-Rewrite—transforming the original questions. This benchmark consists of 1,000 questions, distributes as follows: 220 Numpy questions, 291 Pandas questions, 155 Matplotlib questions, 68 Pytorch questions, 106 Scipy questions, 115 Sklearn 
 questions, 45 Tensorflow questions.  {It is worth mentioning that we opt for a non-default version of the two candidate system prompts: "\textit{Only provide the code completion needed. Don't repeat the context code.}".}
\begin{table}[!t]
  \caption{Performance of \sysname and Popular Small Code LLMs on the DS-1000 Benchmark. DR=Difficult-Rewrite, Mat=Matplotlib.} 
  \centering
  \label{tab:ds1000_eva}
  \scriptsize
  \resizebox{\linewidth}{!}{
  \begin{tabular}{l|cc|ccccccc|cccc|c}
  \toprule
   \multirow{2}{*}{\textbf{Model}} & \multicolumn{2}{c|}{\textbf{Size}} & \multicolumn{7}{c|}{\textbf{Library}} & \multicolumn{4}{c|}{\textbf{Perturbation}} & \multirow{2}{*}\textbf{\textbf{Overall}} \\
   & \#TP & \#AP & Numpy & Pandas & Pytorch & Scipy & Sklearn & TF & Mat & Origin & Surface & Semantic & DR &  \\
   \midrule
   \hline
  Qwen2.5-Coder     & 7B & --  & 49.1 & 29.2 & 39.7 & \textbf{37.7} & 26.1 & 37.8 & \textbf{63.2} & 51.5 & 28.3 & 41.0  & 20.4 & 40.6 \\
  OpenCoder         & 8B & --  & 51.8 & 33.3 & 35.3 & 36.8 & 41.7 & 35.6 & 51.6 & 50.9 & 29.6 & 41.5  & 28.4 & 41.8 \\
  \sysname   & 16.8B & 2.75B & \textbf{56.4} & \textbf{36.1} & \textbf{44.1} & 36.8 & \textbf{47.8} & \textbf{40.0} & 58.1 & \textbf{55.3} & \textbf{40.8} &  \textbf{43.6} & \textbf{29.0} & \textbf{46.1} \\

  \hline
  \bottomrule
  \end{tabular}}
\end{table}

As shown in Table~\ref{tab:ds1000_eva}, overall, \sysname demonstrates the best performance, particularly excelling in the Surface perturbation type, where it significantly outperforms other models. Surface perturbations involve transforming the original problem descriptions while keeping the reference answer unchanged, which highlights \sysname's robust problem understanding and generalization abilities. \sysname also maintains a leading position in the Semantic and Difficult-Rewrite perturbation types.

\textbf{Spider.}
Databases have become integral to various aspects of everyday life, making SQL an essential and widely used tool. Writing SQL queries can be challenging, which has led to a strong demand within tech companies for generating SQL statements from natural language descriptions, i.e. the Text-to-SQL task.
To test our model's Text-to-SQL capabilities, we choose the Spider~\citep{yu2019spider} benchmark. This dataset contains 2,147 problems across four difficulty levels:\textit{easy, medium, hard, extra-hard}, allowing for a comprehensive evaluation of the model's ability to generate SQL queries ranging from simple to highly complex. 
As shown in Table~\ref{tab:spider_eva}, \sysname ranks second, outperforming the DeepSeek-Coder-V2-Lite-Instruct and OpenCoder-8B-Instruct models, but still trailing behind the Qwen2.5-Coder-7B-Instruct model. After a comprehensive comparison and analysis, we believe it is necessary to incorporate text-to-SQL related synthetic data in the pre-training phase and create higher quality text-to-SQL fine-tuning data. This will be one of our next steps.

\begin{table}[!t]
  \caption{Text-to-SQL Performance of \sysname and Comparable Parameter Size Code LLMs on the Spider} 
  \centering
  \label{tab:spider_eva}
  \begin{tabular}{l|cc|ccccc}
  \toprule
   \multirow{2}{*}{\textbf{Model}} & \multicolumn{2}{c|}{\textbf{Size}} &  \multicolumn{5}{c}{\textbf{Spider}} \\
    & \#TP & \#AP  & Easy & Medium & Hard & Extra-Hard & Full \\
  \midrule
  \hline
  CodeQwen1.5     &  7B & --   & 88.5 & 76.3 & 68.6 & 55.90 & 72.9 \\
  DeepSeek-Coder-V2 & 16B & 2.4B & 89.6 & 82.0 & 61.6 & 61.3 & 75.8 \\
  Qwen2.5-Coder     & 7B  & --  & 90.2 & 87.6 & 74.5 & 68.3 & \textbf{82.0} \\
  OpenCoder         & 8B  & -- & 83.4 & 78.9 & 51.8 & 46.2 & 68.1 \\ 
  \sysname   & 16.8B  & 2.75B & 87.7 & 81.2 & 67.6 & 68.1 & 77.5 \\
  \hline
  \bottomrule
  \end{tabular}
\end{table}

\section{Efficiency Analysis}
In this section, we analyze efficiency of \sysname, with Section 5.1 focusing on theoretic comparison between FLOPs and accuracy, and Section 5.2 focusing on practical speed.

\subsection{FLOPs vs Accuracy}

We first investigate the trade-off between accuracy and theoretical computational cost per inference for several compared models. Specifically, we measure the performance of each model in terms of the average scores on 12 benchmarks, as shown in Figure~\ref{fig:merge_performance}, and evaluate their theoretical computational requirements in terms of tera-floating point operations per second (TFLOPs) for a single inference with context length 4096, as illustrated in Figure~\ref{fig:merge_accu_tflops}.

The results indicate that the \sysname achieves the best average score, outperforming both the Qwen2.5-Coder-7B-Instruct and DeepSeek-Coder-V2-Lite-Instruct models. 
Additionally, while the theoretical computational cost per inference for the \sysname is marginally higher than that of the DeepSeek-Coder-V2-Lite-Instruct, it has much less theoretical computational cost than the Qwen2.5-Coder-7B-Instruct and the OpenCoder-8B-Instruct. 
These findings imply that the Ling-Coder-Lite model achieves a better balance between performance and efficiency.

\subsection{Efficiency in Practice}
\begin{figure}[t]
    \centering    \includegraphics[width=0.75\textwidth]{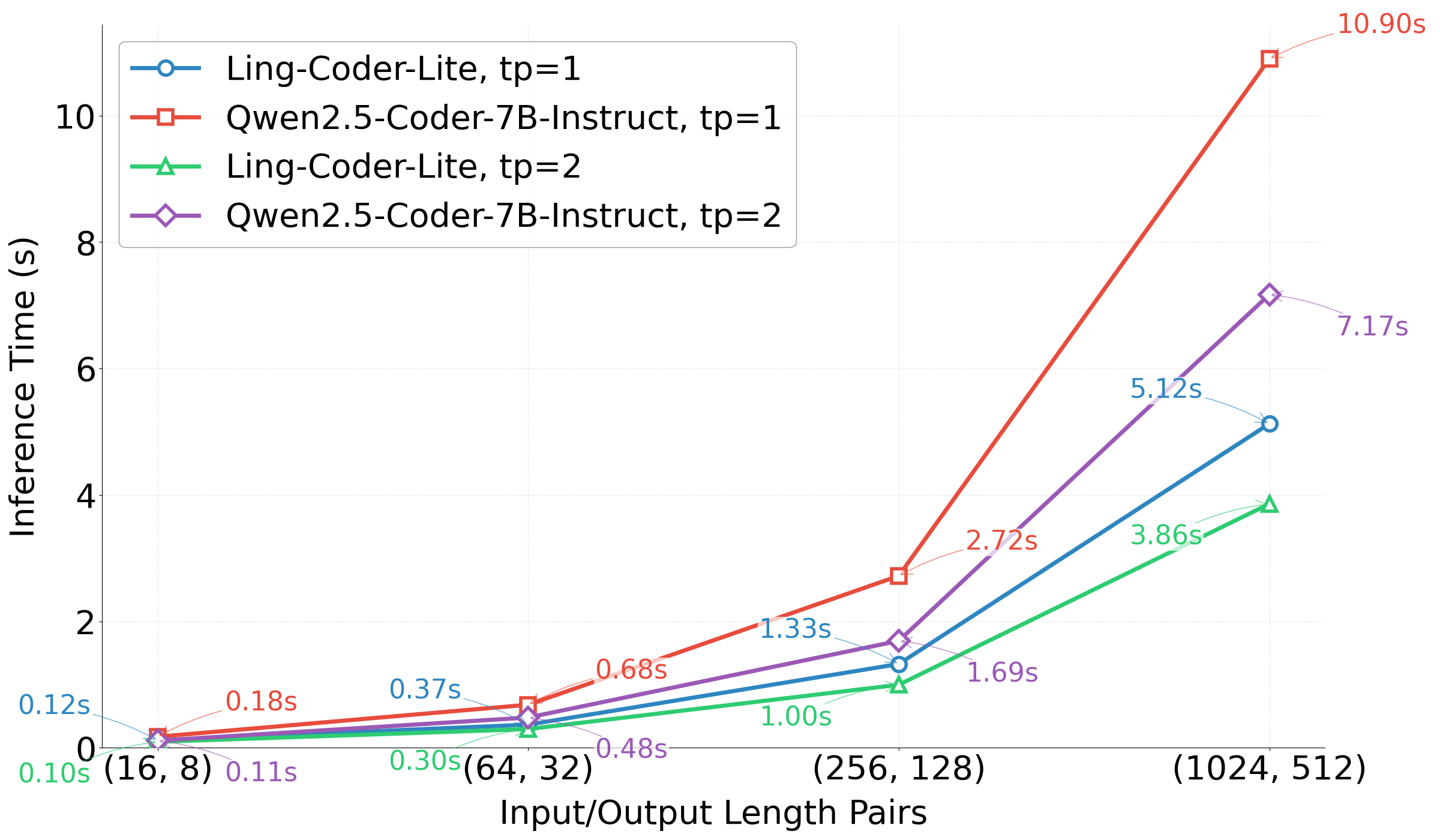}
    \caption{Comparison of Inference Latency between Ling-Coder-Lite and Qwen2.5-Coder-7B-Instruct.}
    \label{fig:inference_time}
\end{figure}

We further conduct a comparative analysis of inference efficiency under practical setting across various input and output lengths between our MoE based Ling-Coder-Lite and the SOTA dense model Qwen2.5-Coder-7B-Instruct. Specifically, we use the vLLM~\citep{kwon2023efficient} inference framework, with models operating in FP16 precision on L20 GPUs. We examine two tensor parallelism configurations: $tp=1$ (single L20 GPU) and $tp=2$ (two L20 GPUs). We set the batch size to 1, and systematically varying the input-output length pairs from shorter to longer sequences. Inference latency performance is assessed using the $benchmark\_latency$~\footnote{\url{https://github.com/vllm-project/vllm/blob/main/benchmarks/benchmark_latency.py}} script provided by vLLM.

As illustrated in Figure~\ref{fig:inference_time}, Ling-Coder-Lite demonstrates significantly superior inference efficiency compared to Qwen2.5-Coder-7B-Instruct. In the $tp=1$ configuration, Ling-Coder-Lite exhibits \pmb{1.5} to \pmb{2} times faster performance than Qwen2.5-Coder-7B-Instruct, and the advantage becomes more pronounced as the length increases. When scaling to $tp=2$, Ling-Coder-Lite achieves an approximately 1.2x speedup, while Qwen2.5-Coder-7B-Instruct achieves a 1.6x improvement over the corresponding cases of $tp=1$. We hypothesize that the relatively modest speedup for Ling-Coder-Lite is attributable to increased inter-machine communication overhead for expert routing in its MoE architecture under tensor parallelism. Nevertheless, in the $tp=2$ configuration, Ling-Coder-Lite still outperforms Qwen2.5-Coder-7B-Instruct by a factor of \pmb{1.1} to \pmb{1.8}, with the performance gap widening for longer sequences.

In the practice, we have an AI-IDE usage internally,
which has more than 10,000 users and helps users generating hundreds of thousands lines of code per day. The service was previously supported by a dense 7B code LLM. When fixing the same latency level requirement for coding completion related tasks ($\le$500ms), 
we realize about 50\% deployment resources reduction of \sysname compared to similar-sized dense model (7B) without experience loss in terms of both accuracy and throughput.

\section{Conclusion \& Future Work}

We introduce Ling-Coder-Lite, an MoE based code LLM that matches state-of-the-art performance on 12 coding benchmarks while offering competitive latency and throughput compared to similar-sized code LLMs. 
We present details on how to build such a coding LLM, including training recipes/strategies for continuous pre-training and post-training, as well as data curation methods for producing high quality code data, code-related data, and synthesizing instruction data. 
To support further research and development in this area, we open-sourced our models and a substantial amount of high-quality data for the annealing and post-training phases.

The future work will focus on further pushing the ultimate Pareto frontier for powerful yet efficient code LLM. We plan to  improve reasoning performance with chain-of-thoughts and add execution feedback for online iterative training in RLHF, enabling to handle more complicated coding tasks in software engineering. 
Meanwhile, we will further enhance the data quality beyond the syntax error, for instance execution correctness check for the level of function methods, files and even repositories.

\section*{Author List}
Note authors are listed in the alphabet order based on their last name. 

\texttt{Wenting Cai, 
Yuchen Cao,
Chaoyu Chen,
Chen Chen,
Siba Chen,
Qing Cui,
Peng Di, 
Junpeng Fang,
Zi Gong, 
Ting Guo,
Zhengyu He,
Yang Huang,
Cong Li,
Jianguo Li, 
Zheng Li, 
Shijie Lian,
BingChang Liu,
Songshan Luo,
Shuo Mao,
Min Shen,
Jian Wu,
Jiaolong Yang,
Wenjie Yang,
Tong Ye,
Hang Yu, 
Wei Zhang,
Zhenduo Zhang,
Hailin Zhao,
Xunjin Zheng,
Jun Zhou
}

\bibliographystyle{plainnat}
\bibliography{ref}

\end{document}